# PatternNet: A Benchmark Dataset for Performance Evaluation of Remote Sensing Image Retrieval

Weixun Zhou[a], Shawn Newsam[b], Congmin Li[a], Zhenfeng Shao[a,*]

[a] State Key Laboratory of Information Engineering in Surveying, Mapping and Remote Sensing, Wuhan University, No.129 Luoyu Road, Wuhan 430079, China

[b] Electrical Engineering and Computer Science, University of California, Merced, CA 95343, USA

**Abstract**

Remote sensing image retrieval (RSIR), which aims to efficiently retrieve data of interest from large collections of remote sensing data, is a fundamental task in remote sensing. Over the past several decades, there has been significant effort to extract powerful feature representations for this task since the retrieval performance depends on the representative strength of the features. Benchmark datasets are also critical for developing, evaluating, and comparing RSIR approaches. Current benchmark datasets are deficient in that 1) they were originally collected for land use/land cover classification and not image retrieval; 2) they are relatively small in terms of the number of classes as well the number of sample images per class; and 3) the retrieval performance has saturated. These limitations have severely restricted the development of novel feature representations for RSIR, particularly the recent deep-learning based features which require large amounts of training data. We therefore present in this paper, a new large-scale remote sensing dataset termed "PatternNet" that was collected specifically for RSIR. PatternNet was collected from high-resolution imagery and contains 38 classes with 800 images per class. We also provide a thorough review of RSIR approaches ranging from traditional handcrafted feature based methods to recent deep learning based ones. We evaluate over 35 methods

---

* Corresponding author: shaozhenfeng@whu.edu.cn

to establish extensive baseline results for future RSIR research using the PatternNet benchmark

# 1. Introduction

Over the past several decades, remote sensing has experienced dramatic changes in the increased spatial resolution of the imagery as well as the increased rate of acquisition. These changes have had profound effects on the way that we use and manage remote sensing images. The increased spatial resolution provides new opportunities for advancing remote sensing image analysis and understanding, making it possible to develop novel approaches that were not possible before. The increased acquisition rate enables us to acquire a considerable volume of remote sensing data on a daily basis. But this has resulted in the significant challenge of how to efficiently manage the large data collections, particularly so that the data of interest can be accessed quickly

Content based image retrieval (CBIR) is a useful technique for the fast retrieval of images of interest from a large-scale dataset [1–4]. Therefore, the remote sensing community has invested significant effort to adapt CBIR to remote sensing images [5–10]. Remote sensing image retrieval (RSIR) is an active and challenging research topic in the field of remote sensing. The remote sensing community has been particularly focused on developing powerful feature extraction methods since retrieval performance depends heavily on the effectiveness of the features [1].

Conventional RSIR methods use low-level visual features to represent the content of the images. These features can be either global or local. Global features are extracted from the whole image, e.g. color (spectral) features [11–14], texture features [5,15–17] and shape features [10]. In contrast to global features, local features like Scale Invariant Feature Transform (SIFT) [18] are extracted from image patches that are centered at interesting points. Local features enjoy several advantages over global ones such as robustness to occlusion as well as invariance to viewing angle and lighting conditions. The remote sensing community has sought to exploit these properties of local features and a number of novel methods have been proposed for image registration [19–24], change detection [25,26], building and urban region detection [27–29], image classification [30] and remote sensing image retrieval [7,31] These local and global features are hand-crafted though. Their development is time consuming and often involves ad-hoc or heuristic design decisions, making them suboptimal for the task at hand.

Deep learning has dramatically advanced the state-of-the-art recently in speech recognition, image classification and object detection [32]. Unlike hand-crafted features, deep learning is capable of discovering intricate structure in large data sets to learn powerful feature representations. It learns the optimal feature representations from the data. Deep learning has been applied to CBIR. Wan *et al.* perform a comprehensive study on deep learning for CBIR with the goal of addressing the fundamental problem of feature representation in CBIR, bridging the semantic gap between the low-level image pixels captured by machines and the high-level semantic concepts perceived by human [33]. Inspired by the great success of deep learning, the remote sensing community has realized the potential of applying deep learning techniques for remote sensing tasks such as scene classification [34–38], object detection [39–41], semantic segmentation [42,43], image super-resolution [44] and image retrieval [45–49]. For RSIR, the deep learning techniques can be roughly divided into unsupervised feature learning and supervised feature learning methods. Generally, supervised feature learning methods, such as convolutional neural networks (CNNs), outperform unsupervised feature learning methods, such as k-means and auto-encoders [50], by a wide margin.

Though the remote sensing community has achieved notable progress in RSIR in recent years, particularly through deep-learning based methods, a comprehensive survey of existing methods applied in a consistent fashion to a single benchmark datasets is lacking. Existing evaluations are incomplete in that they are performed using different performance metrics, on different datasets, and/or under different experimental configurations. There are two fundamental challenges to performing a consistent evaluation. First is the effort and technical challenges of re-implementing existing methods to produce results that can be meaningfully compared. Second is establishing consistent experimental conditions, central to which is having a rich evaluation dataset. The UC Merced dataset [7] was the first publicly available remote sensing evaluation dataset, and it has been used extensively to develop and evaluate RSIR methods [5,17,31,46–49,51,52]. The UC Merced dataset, however, is not very challenging in that it consists of just 21 classes with 100 images in each, all having the same spatial resolution. As a result, retrieval performance has saturated on this dataset. The UC Merced dataset is also too small to develop deep learning based methods for RSIR. Recently, two new large-scale remote sensing datasets, AID [53] and NWPU-RESISC45 [54], have been made publicly available for remote sensing scene classification. These datasets are more challenging and larger than the UC Merced dataset and they contain images at varying spatial resolution. However, the AID and NWPU-RESISC45 datasets are more appropriate for land use/land cover or scene classification than RSIR. We discuss this further in the section on datasets below. In order to stimulate research in RSIR, particularly deep-learning based methods, it is necessary to construct a large-scale dataset appropriate for image retrieval.

In this paper, we first provide a comprehensive review of existing RSIR approaches ranging from traditional handcrafted feature based methods to recently-developed deep-learning feature based methods. We then introduce a large-scale remote sensing image retrieval dataset, named PatternNet. PatternNet provides the remote sensing community with a publicly available benchmark dataset to develop novel algorithms for RSIR. The main contributions of this paper are as follows:

- We provide a comprehensive review of the existing state-of-the-art methods for remote sensing image retrieval (RSIR), ranging from traditional handcrafted feature based methods to recently developed deep learning feature based methods.
- We construct a large-scale remote sensing benchmark dataset, PatternNet, for RSIR. PatternNet is a publicly available, high-resolution dataset which contains more classes and more images than the current RSIR datasets.
- We evaluate existing state-of-the-art methods including handcrafted features and deep learning features on PatternNet under consistent experimental conditions. This provides the literature with extensive baseline results on PatternNet for future research on RSIR.

The rest of this paper is organized as follows. We provide a comprehensive review of existing methods including handcrafted features and deep learning features for RSIR in Section 2. Section 3 reviews several publicly available remote sensing datasets and introduces our large-scale dataset PatternNet. The state-of-the-art methods that are evaluated on PatternNet are described in detail in Section 4. The results and comparisons of these methods are shown in Section 5. Section 6 draws some conclusions.

## 2. Remote Sensing Image Retrieval Methods

The retrieval performance of RSIR methods depends heavily on the effectiveness of the feature representations. Significant effort has therefore been undertaken to develop powerful feature representations over the past few decades. Existing feature representations for RSIR can be generally categorized into two groups, handcrafted features and deep learning features. Note that the two categories are not strictly distinct—hybrid or combinations have also been considered.

*2.1. Handcrafted Feature Based Methods*

*2.1.1. Methods Based on Low-Level Features*

Traditional RSIR methods rely on handcrafted low-level visual features to represent the content of remote sensing images. These includes globally extracted features (global features) and locally extracted features (local features).

Generally, there are three kinds of global features: color (spectral) features [11–14], texture features [5,15–17], and shape features [10]. Color and texture features have been used more widely than shape features for RSIR. Remote sensing images typically have several spectral bands (e.g. multi-spectral imagery) and sometimes even have hundreds of bands (e.g. hyper-spectral imagery) and therefore spectral information is crucial for remote sensing image analysis. Bosilj *et al*. explored both global and local pattern spectral features for geographical image retrieval, and implemented pattern spectra features for the first time with a dense strategy [11]. The performance of the global spectral features as well as its new counterpart were evaluated and compared to state-of-the-art approaches on a benchmark dataset, resulting in the best morphology-based results thus far. Sebai *et al.* proposed multi-scale color component features that improve high resolution satellite images retrieval [12]. The color component features are designed to take simultaneously both color and neighborhood information into consideration, achieving better performance than the existing methods. Color features, however, perform poorly when instances of an object/class vary in spectra or spectra are shared between different objects/class. Texture features have therefore been applied to capture the spatial variation of pixel intensity, and, indeed, they have demonstrated remarkable performance on a range of remote sensing tasks including RSIR. Aptoula explored the potential of recently developed multiscale texture descriptors, the circular covariance histogram and the rotation-invariant point triplets, for the problem of geographic image retrieval, and introduced several new descriptors based on the Fourier power spectrum [5]. These descriptors were shown to outperform the best retrieval scores in spite of their low dimensions. However, most existing texture features are extracted from greyscale images, discarding the useful color information of remote sensing images [55]. Shao *et al.* therefore proposed improved color texture descriptors for RSIR which incorporate discriminative information among color channels [17]. Zhu *et al.* proposed a multi-scale and multi-orientation texture transform spectrum to perform two-level coarse-to-fine rotation- and scale-invariant texture image retrieval [16]. Experiments on a benchmark texture dataset show that the proposed approach captures the primary orientation of the image and generates an informative descriptor. There are other works that focus on combining color and texture features to improve the performance of hyperspectral imagery retrieval [56].

Unlike global features, local features are extracted from image patches centered at interesting points in an image [57,58]. SIFT [18] is one of the most popular local feature descriptors and has been used widely for various remote sensing tasks including scene classification, RSIR, etc. SIFT feature descriptors were compared to 12 features for very high resolution satellite image scene classification in [30]. The results show that SIFT descriptors outperform the other features. Yang *et al.* investigated the use of local invariant features to perform an extensive evaluation of geographic image retrieval on the first publicly available land use/land cover evaluation dataset [7]. The dataset, known as the UC Merced 21-class dataset, has become a widely used benchmark dataset for RSIR and remote sensing scene classification. The local invariant features are compared with several global features, such as simple statistics, color histogram, and texture. The extensive experiments indicate the superiority of local invariant features over global features. In [31], the performance of various image representations for image search problems for geographic image retrieval are investigated. The results demonstrate the suitability of local features for RSIR. Shechtman *et al.* proposed a local self-similarity (SSIM) descriptor [59] to measure the similarity between images or videos based on internal similarities. This descriptor is shown to be efficient and effective for deformable shape retrieval [60], as well as for remote sensing scene classification [30] and multi-sensor remote sensing image matching [61]. Other popular local features include histogram of oriented gradient (HOG) [62] and its variant, descriptor pyramid histogram of oriented gradient (PHOG) [63]. For object detection in remote sensing images, several extensions of HOG have been developed in order to improve detection performance [64–67].

Though local features show better performance than global features on various remote sensing tasks, they are not mutually exclusive but can be complementary. There has been work on combining global and local features to improve performance [68–75].

*2.1.2. Methods Based on Mid-Level Features*

In general, local features like SIFT are of high dimension and numerous, making them impractical for large-scale RSIR. Methods have therefore been developed to transform the local, low-level features into mid-level representations [76] of intermediate complexity through feature encoding techniques such as bag of visual words (BOVW) [77], vector of locally aggregated descriptors (VLAD) [78], and improved fisher kernel (IFK) [79]. BOVW is one of the most popular mid-level features and has been widely used to encode local features into a compact global image representation. BOVW and its variant methods have shown remarkable performance not only in image retrieval [7,31,52,80] but also in remote sensing scene classification [30,81–86]. In [7], BOVW obtained by encoding saliency and grid based SIFT descriptors is evaluated and compared to several global features on a publicly available dataset for geographic image retrieval. The extensive experiments demonstrate the superiority of BOVW over these global features. In [31], BOVW is compared with VLAD and its more compact version, product quantized VLAD (VLAD-PQ) [87], for the purpose of geographic image retrieval from satellite imagery. The results show that VLAD-based representations are more discriminative than BOVW in almost all the land cover classes.

BOVW is not only an image representation but also a framework that can be combined with other features to extract even more powerful representations. For instance, in [52], morphological texture descriptors are combined with the BOVW paradigm in order to extract bag of morphological words for content-based geographic image retrieval. The existing global morphological texture descriptors are adapted to local sub-windows. These local descriptors are then used to form a vocabulary of "visual morphological words" through clustering. In [53], SIFT, local binary pattern (LBP) [88], and color histogram (CH) [89] are combined with BOVW to extract mid-level local features.

Other works focus on improving the BOVW framework in order to achieve better performance. For instance, in [80], an improved BOVW framework is proposed for remote sensing image retrieval in large-scale image databases, which has better performance than the typical BOVW framework and requires less storage cost. Some other works that are based on BOVW can be found in [90–97].

Though BOVW and its variants have achieved remarkable performance on various tasks, the major limitation of such approaches is that the spatial distribution of local features is ignored, which has been proven to be very helpful in improving retrieval performance [98]. Therefore, methods have been proposed to incorporate the spatial arrangement of local features. Cao *et al.* proposed spatial bag of features to encode the geometric information of objects within an image [99] for large scale image retrieval. Compared with BOVW, the spatial bag of feature works well for image retrieval since the spatial information is encoded. Bosch *et al.* proposed a pyramid histogram of visual words (PHOW) [100] descriptor as an image appearance representation based on spatial pyramid matching (SPM) [101]. In [86], the spatial pyramid co-occurrence kernel (SPCK) is proposed for image classification by integrating the absolute and relative spatial information that is ignored in the standard BOVW framework.

*2.2. Deep Learning Feature Based Methods*

Deep learning is a recently developed technique which has dramatically advanced the state-of-the-art in various computer vision tasks including image classification and object detection [32]. CBIR has also benefited from the success of deep learning [33]. As mentioned above, image retrieval performance depends heavily on the effectiveness of the features used to compute image similarity. Deep learning has demonstrated that it is capable of deriving powerful feature representations.

*2.2.1. Unsupervised Feature Learning Based Methods*

Unsupervised feature learning aims to directly learn powerful feature representations from large volumes of unlabeled data. It is therefore attractive for remote sensing since the field has relatively little labeled data compared with many other image analysis areas. In [36], an unsupervised feature learning approach combining SIFT and sparse coding is proposed to learn sparse feature representations for aerial scene classification. Since then a number of unsupervised feature learning approaches have been proposed for various remote sensing applications including remote sensing scene classification [35,37,38] and RSIR [45,46,49,102]. In [49], an unsupervised feature learning framework based on auto-encoders [50] is proposed to learn sparse feature representation for high-resolution remote sensing imagery retrieval and shows better performance than handcrafted BOVW features. In a recent work [46], a novel content-based remote sensing image retrieval approach is proposed via multiple feature representation and collaborative affinity metric fusion. This approach can generate four types of unsupervised features that outperform several handcrafted features on two publicly available datasets. Wang *et al.* developed a novel graph-based learning method for effectively retrieving remote sensing images based on a three-layer framework [45]. This framework integrates the strengths of query expansion and fusion of holistic and local features, achieving remarkable performance on the benchmark dataset.

In contrast to traditional handcrafted features, unsupervised feature learning based methods directly learn powerful feature representations from the data for RSIR. The performance improvement, however, has been limited. This is because the unsupervised feature learning methods mentioned above are often based on shallow networks (e.g. the three-layer auto-encoder in [49]) which cannot learn higher-level information. It is therefore worth investigating deeper networks in order to extract more discriminative features for RSIR.

*2.2.2. Convolutional Neural Networks Based Methods*

Convolutional neural networks (CNNs) have proven to be the most successful deep learning approach to image analysis based on their remarkable performance on the large-scale benchmark dataset ImageNet [103]. CNNs can learn high-level feature representations that are more discriminative than unsupervised features via a hierarchical architecture consisting of convolutional, pooling, and fully-connected layers. However, large numbers of labeled images are needed to train effective CNNs. Transfer learning is often used to remedy this by treating the CNNs pre-trained on ImageNet as feature extractors, possibly fine-tuning the pre-trained CNNs on the target dataset to learn domain-specific features. This is very helpful for some domains (e.g. remote sensing field) where large-scale publicly available datasets are lacking. In [104], the generalization power of deep features extracted by CNNs is investigated by transferring deep features from everyday objects to aerial and remote sensing domains. Experiments demonstrate that transfer learning is an effective approach for cross-domain tasks.

Currently, CNNs have been widely used for various retrieval tasks ranging from computer vision to remote sensing [47,48,105–114]. In [47], Zhou *et al.* proposed two effective schemes to investigate how to extract powerful feature representations based on CNNs for high-resolution remote sensing imagery retrieval. In the first scheme, the convolutional and full-connected layers of pre-trained CNNs are regarded as feature extractors, while in the second scheme, a novel CNN is proposed to learn low dimensional features from limited labeled images. The two schemes, and in particular the novel low dimensional CNNs, achieve state-of-the-art performance on several evaluation datasets.

In [48], an extensive evaluation of visual descriptors including handcrafted global and local features as well as CNN features is conducted for content-based retrieval of remote sensing images. The results demonstrate that CNN-based features usually outperform handcrafted features except for remote sensing images that have more heterogeneous content.

It should be noted that although deep learning feature based methods can directly learn powerful feature representations and often outperform handcrafted feature based methods for RSIR, they still have several limitations. A large number of samples are needed to train effective deep learning models, particularly supervised models like CNNs which requires large amounts of labeled data. However, there is a lack of such datasets in remote sensing. The other limitation is that "tricks" are often necessary to speed up the training and to achieve satisfactory performance. This makes it difficult and time consuming to determine the optimal model for the target task.

**3. PatternNet: A Large-Scale Dataset for Remote Sensing Image Retrieval**

This section first reviews several publicly available remote sensing datasets and then introduces the proposed large-scale high-resolution dataset (PatternNet) for RSIR.

*3.1. The Existing Remote Sensing Datasets*

**UC Merced dataset** (http://vision.ucmerced.edu/datasets/landuse.html). The UC Merced dataset (UCMD) [7] is a land use/land cover dataset which contains 100 images of the following 21 classes: agricultural, airplane, baseball diamond, beach, buildings, chaparral, dense residential, forest, freeway, golf course, harbor, intersection, medium density residential, mobile home park, overpass, parking lot, river, runway, sparse residential, storage tanks and tennis courts. Each image measures 256 × 256 pixels. The images are cropped from large aerial images downloaded from the United States

Geological Survey (USGS) and the spatial resolution is around 0.3m. The UCMD dataset has several highly overlapping classes (*i.e.* sparse residential, medium residential and dense residential), which makes it a challenging dataset. The UCMD dataset has been a benchmark dataset for RSIR.

**WHU-RS19 dataset** (http://dsp.whu.edu.cn/cn/staff/yw/HRSscene.html). The WHU-RS19 remote sensing dataset (RSD) [115] is manually collected from Google Earth Imagery and labeled into 19 classes: airport, beach, bridge, commercial area, desert, farmland, football field, forest, industrial area, meadow, mountain, park, parking, pond, port, railway station, residential area, river, and viaduct. The dataset consists of a total of 1,005 images and each image has the size of 600 × 600 pixels. The images in RSD have a wide range of spatial resolutions which are up to 0.5m.

**RSSCN7 dataset** (https://www.dropbox.com/s/j80iv1a0mvhonsa/RSSCN7.zip?dl=0). The RSSCN7 dataset [116] is sampled on four different scale levels from Google Earth imagery and consists of 7 classes: grassland, forest, farmland, parking lot, residential region, industrial region, river, and lake. There are 400 images in each class and each image has size of 400 × 400 pixels.

**Aerial image dataset** (http://www.lmars.whu.edu.cn/xia/AID-project.html). The aerial image dataset (AID) [53] is a recently released large-scale dataset, which is collected with the goal of advancing the state-of-the-art in scene classification of remote sensing images. It is notably larger than the three datasets mentioned above and contains 30 classes: airport, bare land, baseball field, beach, bridge, center, church, commercial, dense residential, desert, farmland, forest, industrial, meadow, medium residential, mountain, park, parking, playground, pond, port, railway station, resort, river, school, sparse residential, square, stadium, storage tanks, and viaduct. There are a total of 10,000 images in the AID dataset and each class has 220~420 images of size 600 × 600 pixels. The spatial resolution of this dataset varies greatly between approximately 0.5 to 8 m.

**NWPU-RESISC45 dataset** (https://1drv.ms/u/s!AmgKYzARBl5ca3HNaHIlzp_IXjs). The NWPU-RESISC45 dataset [54] is currently the largest publicly available benchmark dataset for remote sensing scene classification. It was also recently released. It is constructed by first investigating all scene classes of the existing datasets and then selecting a list of 45 representative scene classes: airplane, airport, baseball diamond, basketball court, beach, bridge, chaparral, church, circular farmland, cloud, commercial area, dense residential, desert, forest, freeway, golf course, ground track field, harbor, industrial area, intersection, island, lake, meadow, medium residential, mobile home park, mountain, overpass, palace, parking lot, railway, railway station, rectangular farmland, river, roundabout, runway, sea ice, ship, snow berg, sparse residential, stadium, storage tank, tennis court, terrace, thermal power station, and wetland. Each class has 700 images of size 256 × 256 pixels and the spatial resolution of the images in each class varies from about 0.2 to 30m.

Though there are five publicly available remote sensing datasets, the UCMD dataset is the one that has been used the most widely as a benchmark for RSIR. However, the UCMD dataset is a small-scale dataset with only 21 classes with 100 images in each class, and all the images have the same spatial resolution. This is not sufficient for developing novel approaches to RSIR based on deep learning. Further, the retrieval performance on this dataset has saturated.

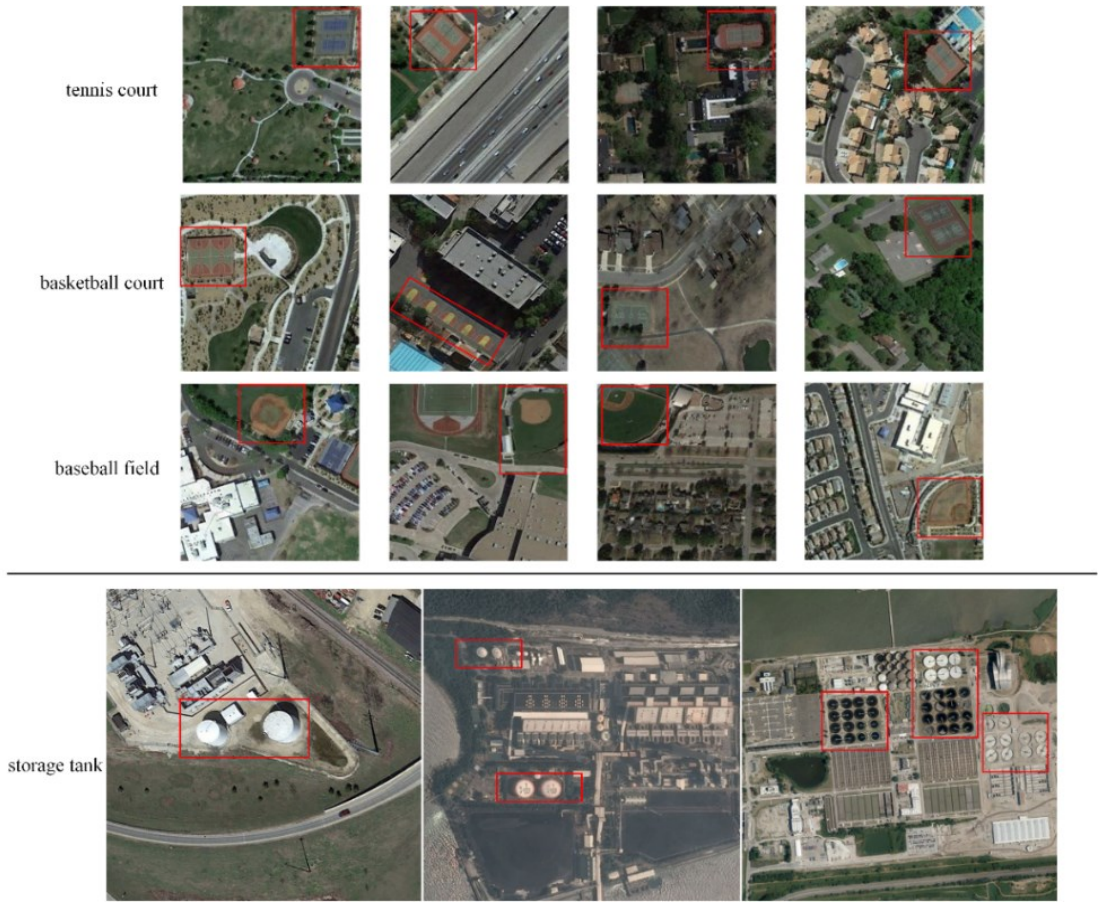

**Figure 1**. Some example images from the NWPU-RESISC45 (top row) and the AID (bottom row) datasets.

AID and in particular NWPU-RESISC45 are two large-scale datasets compared to the UCMD and the RSSCN7 datasets. However, they are collected for remote sensing scene classification and have greatly varying spatial resolutions ranging from low to high-resolution. Significantly, many images contain a large amount of background which is not appropriate for RSIR. Figure 1 shows example images from the NWPU-RESISC45 and AID datasets. For the NWPU-RESISC45 dataset, we can see the class of interest only covers a small portion of the image and is located in the corner. This kind of imagery is not appropriate for developing and evaluating image retrieval methods as the background is unlikely to be relevant for a query and yet will dominate the image representation. Take the fourth tennis court image for example. A query using this image would be more likely to return residential areas than tennis courts. The images in the AID dataset also contain a large amount of background. The AID images are also larger at 600 x 600 pixels and cover more area. This makes them more likely to be mislabeled. Take the third AID image for example. It is labeled as storage tank but the more appropriate label is wastewater treatment plant.

*3.2. The Proposed Large-Scale Dataset for Remote Sensing Image Retrieval*

**The large-scale dataset PatternNet**. PatternNet[1] is a large-scale high-resolution remote sensing dataset collected for RSIR. It contains 38 classes: airplane, baseball field, basketball court, beach, bridge, cemetery, chaparral, Christmas tree farm, closed road, coastal mansion, crosswalk, dense residential, ferry terminal, football field, forest, freeway, golf course, harbor, intersection, mobile home park, nursing home, oil gas field, oil well, overpass, parking lot, parking space, railway, river, runway, runway marking, shipping yard, solar panel, sparse residential, storage tank, swimming pool, tennis court, transformer station and wastewater treatment plant. There are a total of 800 images of size 256 × 256 pixels in each class. The dataset name "PatternNet" is inspired by the project TerraPattern [117], an open-source tool for discovering "patterns of interest" in unlabeled satellite imagery which provides an open-ended interface for visual query-by-example.

The images in PatternNet are collected from Google Earth imagery or via the Google Map API for US cities. Table 1 shows the details of PatternNet. Similar to the AID and the NWPU-RESISC45 datasets, PatternNet contains images with varying resolution; but its images generally have much higher resolution. The highest spatial resolution is around 0. 062m and the lowest spatial resolution is around 4.693m.

---

[1] PatternNet is available at https://sites.google.com/view/zhouwx/dataset

**Table 1.** The details of PatternNet dataset. "GMA" means the images are collected using Google Map API and "GE" means the images are collected from Google Earth imagery.

| Class | Resolution (meter/pixel) | | Source | |
|---|---|---|---|---|
| | GMA | GE | GMA | GE |
| airplane | N/A | 0.217 | No | Yes |
| baseball field | 0.233~0.293 | 0.124 | Yes | Yes |
| basketball court | 0.116~0.146 | 0.161 | Yes | Yes |
| beach | N/A | 0.158 | No | Yes |
| bridge | 0.465~0.586 | 0.466 | Yes | Yes |
| cemetery | 0.233~0.293 | N/A | Yes | No |
| chaparral | 0.233~0.293 | N/A | Yes | No |
| Christmas tree farm | N/A | 0.124 | No | Yes |
| closed road | 0.233~0.293 | 0.217 | Yes | Yes |
| coastal mansion | 0.233~0.293 | N/A | Yes | No |
| crosswalk | 0.233~0.293 | N/A | Yes | No |
| dense residential | 0.233~0.293 | N/A | Yes | No |
| ferry terminal | 0.465~0.586 | 0.311 | Yes | Yes |
| football field | 0.931~1.173 | 0.817 | Yes | Yes |
| forest | 0.233~0.293 | N/A | Yes | No |
| freeway | N/A | 0.311 | No | Yes |
| golf course | 0.233~0.293 | 0.233 | Yes | Yes |
| harbor | 0.233~0.293 | N/A | Yes | No |
| intersection | 0.465~0.586 | N/A | Yes | No |
| mobile home park | N/A | 0.248 | No | Yes |
| nursing home | 0.465~0.586 | N/A | Yes | No |
| oil gas field | 3.726~4.693 | N/A | Yes | No |
| oil well | N/A | 0.062 | No | Yes |
| overpass | N/A | 0.466 | No | Yes |
| parking lot | 0.233~0.293 | N/A | Yes | No |
| parking space | 0.116~0.146 | 0.102 | Yes | Yes |
| railway | 0.233~0.293 | N/A | Yes | No |
| river | 0.931~1.173 | N/A | Yes | No |
| runway | 0.465~0.586 | N/A | Yes | No |
| runway marking | 0.233~0.293 | N/A | Yes | No |
| shipping yard | 0.233~0.293 | N/A | Yes | No |
| solar panel | 0.233~0.293 | N/A | Yes | No |
| sparse residential | 0.233~0.293 | N/A | Yes | No |
| storage tank | 0.465~0.586 | N/A | Yes | No |
| swimming pool | 0.116~0.146 | N/A | Yes | No |
| tennis court | 0.116~0.146 | 0.158 | Yes | Yes |
| transformer station | 0.233~0.293 | N/A | Yes | No |
| wastewater treatment plant | 0.233~0.293 | 0.124/0.189/0.248 | Yes | Yes |

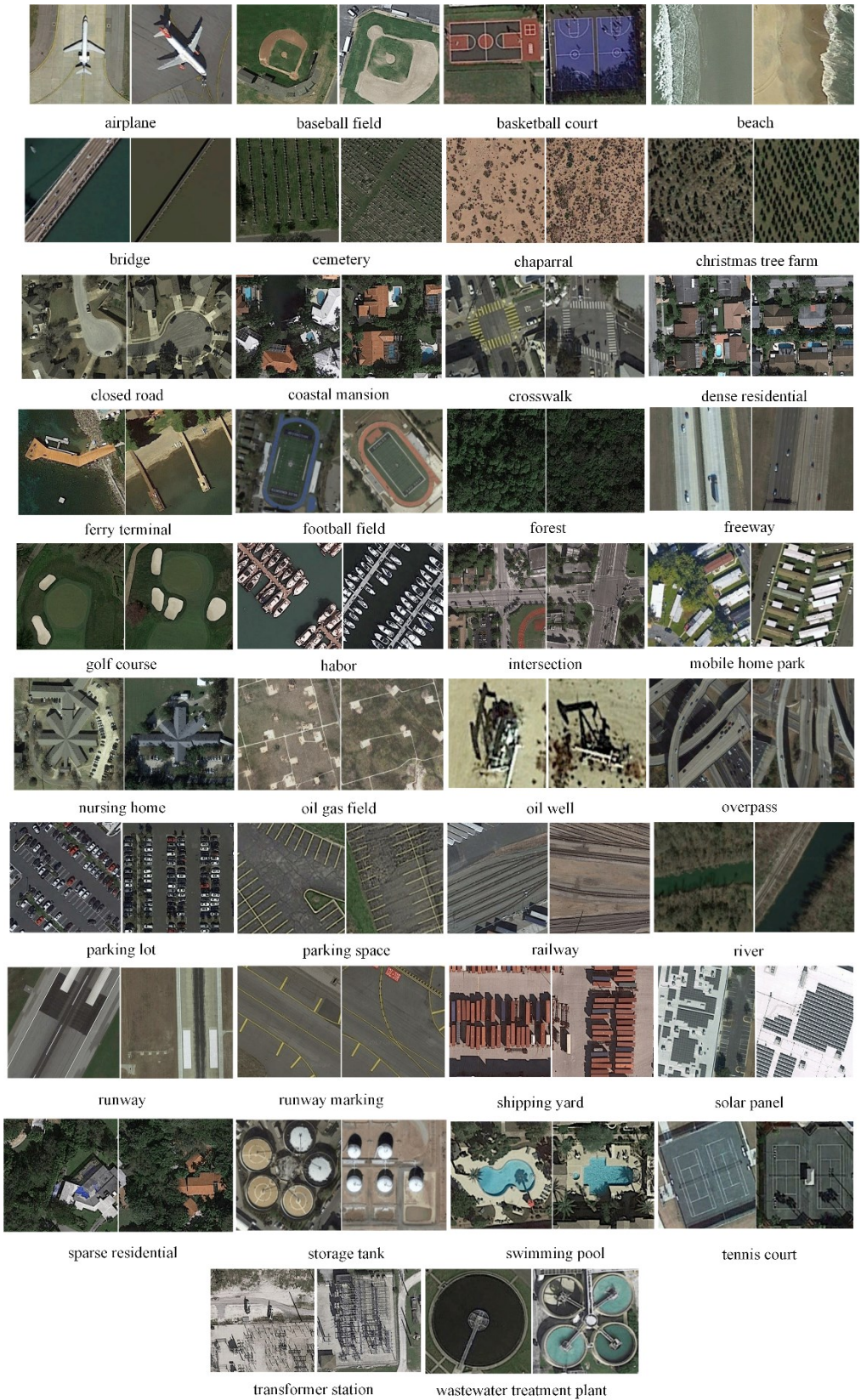

**Figure 2**. Two example images in each class from the PatternNet dataset.

Figure 2 shows example images from the PatternNet dataset. Note that the class of interest covers most of the interest—there is very little background. In particular, see the baseball field, basketball court, closed road, storage tank, and tennis court classes. In summary, our proposed PatternNet dataset has the following notable characteristics.

- Large scale. The PatternNet dataset is the largest publicly available remote sensing dataset collected specifically for RSIR. It is meant to serve as an alternate to UCMD to advance the state-of-the-art in RSIR, particularly deep learning based approaches which require large amounts of labeled training data.

- High resolution. The AID and NWPU-RESISC45 datasets have spatial resolutions ranging from 0.5m to 8m and from 0.2m to 30m respectively. Many images thus cover a large area and contain a large amount of background which is not appropriate for RSIR. In contrast, PatternNet has a higher spatial resolution so that the classes of interest constitute a larger portion of the image.

- High intra-class diversity and inter-class similarity. The lack of high variation and diversity in some datasets has resulted in saturated performance. In contrast, PatternNet was collected from a large number of US cities at varying spatial resolution. PatternNet has both high intra-class diversity (see wastewater treatment plant in Figure 2) and high inter-class similarity (see basketball court and tennis court in Figure 2).

**4. Baseline Methods**

In this section, a large number of the state-of-the-art methods mentioned in the introduction, including those based on handcrafted and deep learning features, are evaluated on the proposed PatternNet dataset.

*4.1. Handcrafted Feature Based Methods*

*4.1.1. Low-Level Handcrafted features*

For low-level visual features, we choose several global features, simple statistics, color histogram [89], Gabor texture [118] and GIST [119], as well as several local features, LBP [88], and PHOG [63].

- **Simple statistics**. Simple statistics is a 2-D feature vector which consists of the mean and standard deviation of the grayscale image pixel intensities.
- **Color histogram**. Color histograms are used to represent the spectral information of remote sensing images. In our experiments, color histograms are extracted by quantizing each channel of the RGB color space into 32 bins and concatenating the three histograms to obtain a 96-D histogram.
- **Gabor texture**. The Gabor filter used in [17] is used to extract Gabor texture features at five scales and eight orientations The same parameters are used except for the size of Gabor filter window which is set to 32 × 32 pixels in our experiments.
- **GIST feature**. GIST is used to represent the dominant spatial structure of an image based on a spatial envelope model. It is widely used for the recognition of real world scenes. The default parameters are used to extract 512-D feature vectors.
- **Local binary pattern (LBP)**. LBP is used to extract local texture information. In our implementation, an 8 pixel circular neighborhood of radius 1 is used to extract a 10-D uniform rotation invariant histogram.
- **Pyramid of histogram of oriented gradients (PHOG)**. PHOG is an extension of HOG which combines the spatial pyramid matching kernel (SPM) [101] to represent the spatial layout of local image shape. The original implementation with the default parameters is used to extract 680-D feature vectors.

*4.1.2. Mid-Level Handcrafted features*

Mid-level features are often extracted by encoding low-level local features like SIFT into global feature representations. In our experiments, three feature encoding approaches, BOVW [77], VLAD [78] and IFK [79], are used to aggregate SIFT descriptors into mid-level feature representations.

- **Bag of visual words (BOVW)**. BOVW is one of the most popular feature descriptors in the last decade. SIFT descriptors are first extracted to represent local image patches, then these descriptors are used to learn a dictionary (also known as codebook or vocabulary of visual words) by k-means clustering. Once the dictionary is constructed, the descriptors of each image can be quantized into the visual words to obtain the global histogram.

- **Vector of locally aggregated descriptors (VLAD)**. VLAD is based on fisher kernels to compute the descriptor. The local descriptors are clustered to construct a dictionary. The VLAD representation is then formed by aggregating the difference vectors between the local descriptors and the visual words. The VLAD representation is a $KD$-D feature vector, where $K$ is the size of dictionary and $D$ is the dimension of local descriptor (e.g. 128 for the SIFT descriptor).

- **Improved fisher kernel (IFK)**. IFK uses Gaussian mixture models to encode local feature descriptors. The IFK representation is formed by concatenating the partial derivatives of the mean and variance of the Gaussian functions. The IFK representation is a $2KD$-D feature vector, where $K$ is the size of dictionary and D is the dimension of the local descriptor.

*4.2. Deep Learning Feature Based Methods*

*4.2.1. Unsupervised Feature Learning Based Methods*

In contrast to handcrafted low-level and mid-level features, unsupervised feature learning methods can directly learn powerful feature representations from unlabeled images. In our experiments, the unsupervised feature learning method (UFL) proposed for high-resolution remote sensing image retrieval in [49] is evaluated on the PatternNet dataset.

UFL is an unsupervised feature learning framework for image retrieval consisting of the four steps shown in Figure 3: (1) local feature extraction, (2) unsupervised feature learning, (3) feature encoding and (4) sparse feature extraction and pooling. The local features extracted from the training images are first fed into an auto-encoder network for unsupervised feature learning. Once trained, the auto-encoder network is used to encode the local feature descriptors to obtain the learned feature set. The final feature representation is then generated by pooling the learned feature descriptors into a global feature vector. We refer the reader to [49] for more details.

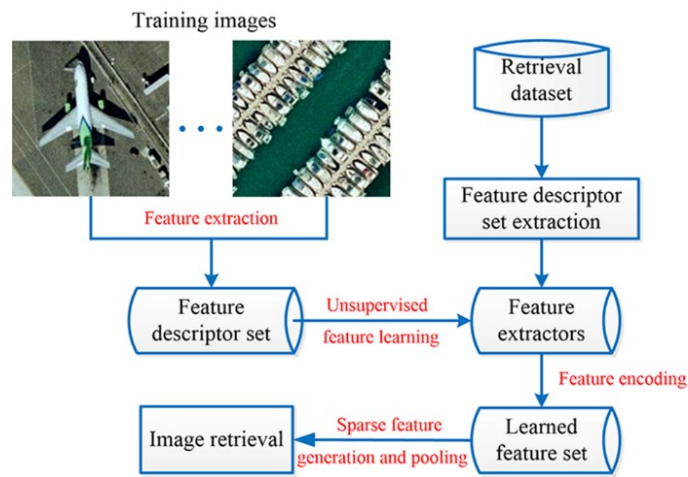

**Figure 3**. The flowchart of the unsupervised feature learning method (UFL). The figure is adapted from previous work [49].

*4.2.2. Convolutional Neural Networks (CNNs)*

Convolutional neural networks (CNNs) are perhaps the most successful deep learning method for image analysis. In contrast to the shallow unsupervised feature learning networks, CNNs are often much deeper and are trained with a large number of labeled images. CNNs thus tend to generate more discriminative features than unsupervised feature learning methods. The basic building blocks of a CNN model consist of various layers including convolutional layers, pooling layers, and fully-connected layers. In general, each convolutional layer has a fixed number of filters (also called kernels or weights) which output the same number of feature maps by sliding the filters over feature maps of the previous layer. The pooling layers are used to reduce the size of the feature maps via sum or max pooling and usually follow the convolutional layers. The fully-connected layers are essentially classifiers that follow the convolutional and pooling layers.

Though CNNs have achieved remarkable performance on various tasks including RSIR, their main limitation is that they require a large number of labeled samples for training. In addition, the features extracted from the fully-connected layers are usually 4096-D feature vectors, which are large and present computational and storage challenges for large-scale RSIR. We therefore evaluate a recent CNN architecture that learns low dimensional features from limited labeled images [47]. The low dimensional CNN (LDCNN) is based on several convolutional layers and a three-layer perceptron. It has fewer parameters than the existing pre-trained CNNs and is therefore much more efficient to train.

In our experiments, some existing pre-trained CNNs as well as the LDCNN are evaluated on the PatternNet dataset.

**The pre-trained CNNs.** Several pre-trained CNNs including the baseline model AlexNet [120], the Caffe (Convolutional Architecture for Fast Feature Embedding) reference model (CaffeRef) [121], the VGG network including VGGF, VGGM and

VGGS [122], the VGG-VD network including VGG-VD16 and VGG-VD19 [123], and recently developed much deeper models, GoogLeNet [124] and Residual networks (ResNet) [125], are used to extract features for RSIR. The framework MatConvNet [126] is used for CNN feature extraction.

- **AlexNet**. AlexNet is regarded as a baseline CNN model. It achieved the best performance in the 2012 ImageNet Large Scale Visual Recognition Challenge (ILSVRC-2012). AlexNet contains five convolutional layers followed by three fully-connected layers.

- **CaffeRef**. CaffeRef can be regarded as a minor variation of AlexNet. The main differences lie in the order of the pooling and normalization layers as well as data augmentation strategy. AlexNet and CaffeRef often achieve similar performance on the same task.

- **VGG**. The VGG network family includes three different CNNs, VGGF, VGGM, and VGGS. These three CNNs are proposed to explore the tradeoff between speed and accuracy. They have similar architecture except for the number and sizes of the filters in the convolutional layers.

- **VGG-VD**. In contrast to VGGF, VGGM and VGGS, VGG-VD is a very deep CNN network. There are two VGG-VD networks, namely VD16 (16 weight layers including 13 convolutional layers and 3 fully-connected layers) and VD19 (19 weight layers including 16 convolutional layers and 3 fully-connected layers). These two CNNs are designed to explore the effect of network depth

- **GoogLeNet**. GoogLeNet is a representative CNN architecture which achieved state-of-the-art performance in the 2014 ImageNet Large Scale Visual Recognition Challenge (ILSVRC-2014). The architecture of GoogLeNet is designed based on

"inception modules". The introduction of inception modules in the architecture gives GoogLeNet two advantages: (1) the spatial information is maintained by using filters with different sizes in the same layer, and (2) the number of parameters is reduced even though GoogLeNet has more than 50 convolutional layers distributed in the Inception modules.

- **ResNet**. Residual networks (ResNet) permit the training of networks that are substantially deeper than those used previously. The layers of ResNet are explicitly reformulated as learning residual functions with reference to the layer inputs. ResNet achieved the best performance in the 2015 ImageNet Large Scale Visual Recognition Challenge (ILSVRC-2015). In our experiments, three ResNet networks are evaluated, ResNet-50 (50-layer ResNet), ResNet-101 (101-layer ResNet), and ResNet-152 (152-layer ResNet).

The above, pre-trained CNNs are used as feature extractors in the experiments. 4096-D features are extracted from the first and second fully-connected layers of the CNNs except for GoogLeNet and ResNet. For GoogLeNet, we extract features from the last pooling layer to generate 1024-D features. In the case of the three ResNet networks, 2048-D features are extracted from the fifth pooling layers.

**The low dimensional CNN (LDCNN).** LDCNN learns low dimensional features from limited training images from scratch. Its low dimension makes it particularly attractive for large-scale RSIR. Figure 4 shows the flowchart of LDCNN. It consists of five convolutional layers and an mlpconv layer (three-layer perceptron). The mlpconv layer is the basic block of the popular CNN architecture Network in Network (NIN) [127] and is used to learn more abstract high-level features that are nonlinearly separable. The global average pooling layer is used to compute the average of each feature map from the previous layer, leading to an *n*-dimensional feature vector (*n* is the number of image classes). We refer the reader to [47] for more details.

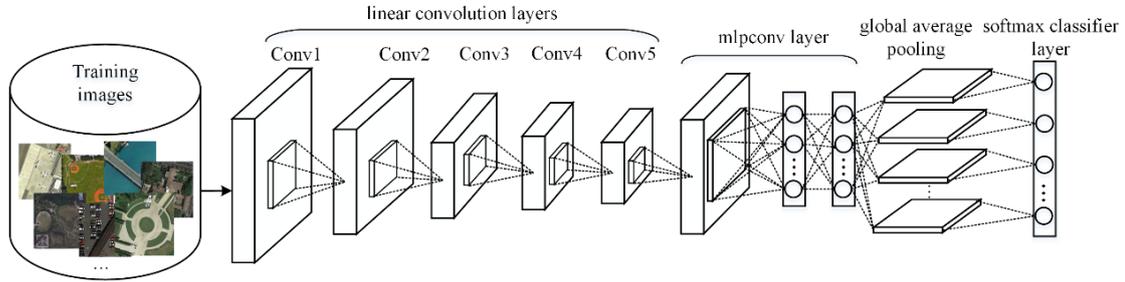

**Figure 4**. The flowchart of the proposed low dimensional CNN (LDCNN). The figure is adapted from previous work [47].

## 5. Experiments and Results

In this section, we evaluate the handcrafted and deep learning feature based methods described in section 4 for RSIR using the proposed PatternNet dataset.

*5.1. Experimental Setup*

The input images of CNNs should have fixed dimensions. Therefore, the images in the PatternNet dataset are resized to 227 × 227 pixels for AlexNet and CaffeRef and to 224 × 224 pixels for the other CNNs. In addition, average images provided by the pre-trained CNNs are subtracted from the resized images. Recent work [47] demonstrates that the activation function, the element-wise rectified linear units (ReLU), has an effect on the performance of the features extracted from the fully-connected layers. In particular, the features extracted from the first fully-connected layer (Fc1 feature) achieve better performance without the use of ReLU, while in the case of the features extracted from the second fully-connected layer (Fc2 feature), better performance results with the use of ReLU. Therefore, in our experiments, Fc1 features are extracted without ReLU and Fc2 features are extracted with ReLU.

With respect to LDCNN, the weights of the five convolutional layers are transferred from VGGF and are also kept fixed during training in order to speed up training. The weights of the mlpconv layer are initialized from a Gaussian distribution (with a mean of 0 and a standard deviation of 0.01). We randomly select 80% of the images from each class of PatternNet as the training set and the remaining 20% of the images are used for retrieval performance evaluation.

For the three mid-level features (*i.e.* BOVW, VLAD and IFK), the dictionary is constructed by aggregating the 128-D SIFT descriptors extracted at the salient points within the image. The dictionary sizes of VLAD and IFK are set to 64 based on the results in [53]. For BOVW, a set of dictionary sizes (*i.e.* 64, 128, 256, 512, 1024, 2048, and 4096) are evaluated. For the unsupervised feature learning method (UFL), the number of neural units in the hidden layer is set to 400, 600 and 800, and the sparsity value is set to 0.4 to generate sparse features,

We empirically select $L1$ as the distance function to computed image similarity for the histogram features including color histogram, BOVW and UFL, and select L2 as the distance function for the remaining features including simple statistics, Gabor texture, GIST, LBP, HOG, PHOG and the CNNs. All the features are $L2$ normalized before the similarity measure is applied. Four commonly used performance metrics, average normalized modified retrieval rank (ANMRR), mean average precision (mAP), precision at $k$ ($P@k$ where $k$ is the number of retrieved images), and precision-recall (PR) curves, are used to evaluate the retrieval performance. In the following experiments, each image is taken as a query image, which means the ANMRR, mAP, and $P@k$ are the averaged values over all the queries.

## 5.2. Experimental Results

### 5.2.1. Results of Handcrafted Low-Level Features

Table 2 shows the performance of the handcrafted low-level features including simple statistics, color histogram, Gabor texture, GIST, LBP, and PHOG measured using ANMRR, mAP and *P@k* (*k*=5, 10, 50, 100, 1000). We can see that Gabor texture features achieve the best performance and simple statistics features achieve the worst performance. In addition, it is shown that the performance decreases as the number of retrieved images increases (the value of *k*). Figure 5 shows the results of these handcrafted features for each class. Simple statistics and PHOG perform worse than the other features for most of the classes in the PatternNet dataset.

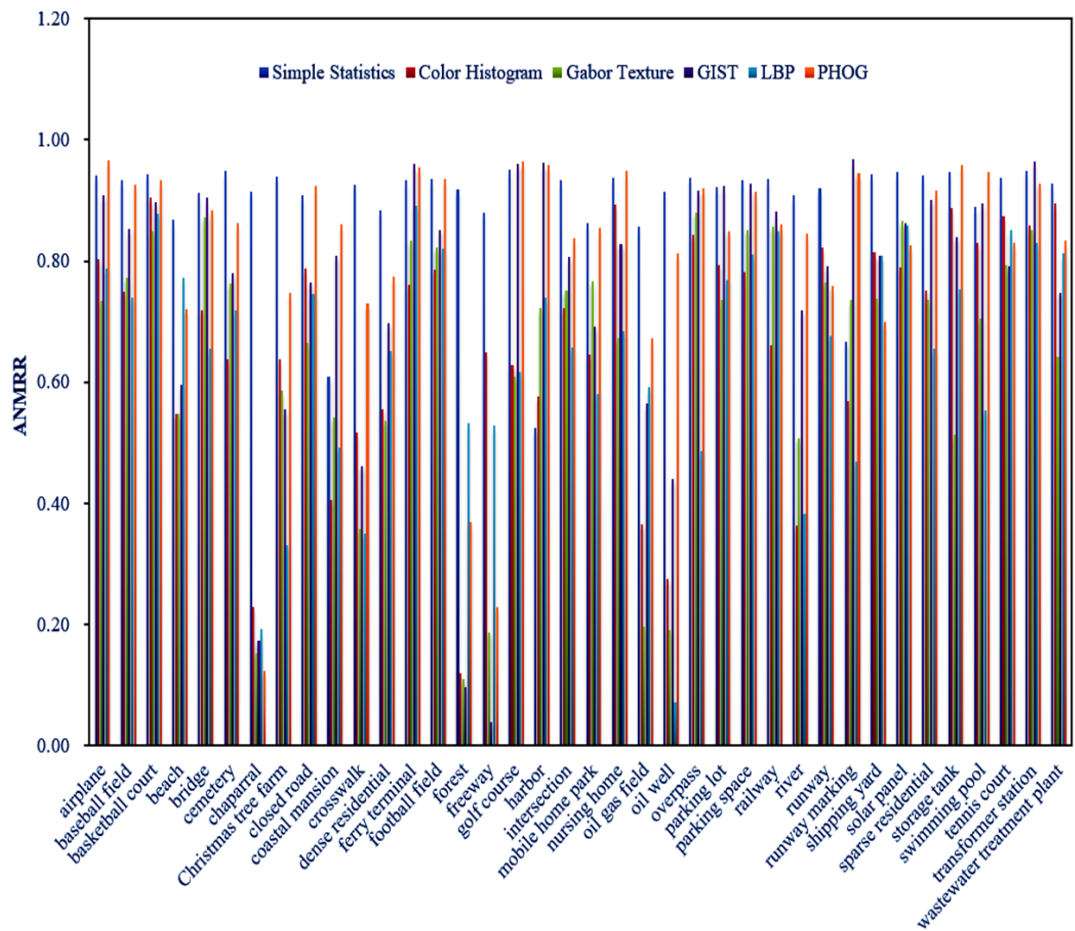

**Figure 5**. The results of low-level features for each class in the PatternNet dataset.

**Table 2.** The results of the handcrafted low-level features on PatternNet. For ANMRR, lower values indicate better performance, while for mAP and *P@k,* larger is better.

| Features | ANMRR | mAP | P@5 | P@10 | P@50 | P@100 | P@1000 |
|---|---|---|---|---|---|---|---|
| Simple Statistics | 0.8968 | 0.0662 | 0.0739 | 0.0741 | 0.0739 | 0.0738 | 0.0701 |
| Color Histogram | 0.6697 | 0.2510 | 0.7475 | 0.7032 | 0.5733 | 0.5062 | 0.2349 |
| Gabor Texture | **0.6422** | **0.2769** | **0.8021** | **0.7631** | **0.6393** | **0.5674** | **0.2556** |
| GIST | 0.7511 | 0.2001 | 0.6429 | 0.5957 | 0.4645 | 0.4013 | 0.1773 |
| LBP | 0.6470 | 0.2583 | 0.6358 | 0.6027 | 0.5115 | 0.4646 | 0.2505 |
| PHOG | 0.8162 | 0.1312 | 0.4852 | 0.4430 | 0.3376 | 0.2903 | 0.1295 |

**Table 3.** The results of the handcrafted mid-level features on PatternNet. For ANMRR, lower values indicate better performance, while for mAP and *P@k,* larger is better. "BOVW-K" means the BOVW extracted with a dictionary size of K.

| Features | ANMRR | mAP | P@5 | P@10 | P@50 | P@100 | P@1000 |
|---|---|---|---|---|---|---|---|
| BOVW-64 | 0.6593 | 0.2536 | 0.5418 | 0.5158 | 0.4506 | 0.4172 | 0.2430 |
| BOVW-128 | 0.6393 | 0.2729 | 0.5853 | 0.5564 | 0.4855 | 0.4489 | 0.2583 |
| BOVW-256 | 0.6573 | 0.2613 | 0.5819 | 0.5498 | 0.4725 | 0.4323 | 0.2450 |
| BOVW-512 | 0.7696 | 0.1781 | 0.3974 | 0.3596 | 0.2721 | 0.2323 | 0.1638 |
| BOVW-1024 | 0.8604 | 0.1111 | 0.2068 | 0.1773 | 0.1213 | 0.1014 | 0.0973 |
| BOVW-2048 | 0.9020 | 0.0820 | 0.1425 | 0.1205 | 0.0782 | 0.0639 | 0.0676 |
| BOVW-4096 | 0.9231 | 0.0667 | 0.0938 | 0.0795 | 0.0565 | 0.0471 | 0.0525 |
| VLAD | **0.5686** | **0.3367** | **0.6466** | **0.6204** | **0.5620** | **0.5318** | **0.3124** |
| IFK | 0.6016 | 0.3093 | 0.6310 | 0.6049 | 0.5436 | 0.5114 | 0.2874 |

*5.2.2. Results of Handcrafted Mid-Level Features*

The results of the mid-level features are shown in Table 3. For the BOVW features, a set of dictionary sizes (64, 128, 256, 512, 1024, 2048, 4096) are investigated. We can see

BOVW with a dictionary size of 128 achieves better performance than BOVW with the other dictionary sizes. In contrast to BOVW, the higher dimensional features VLAD and IFK achieve about 7% and 4% improvement respectively in terms of ANMRR value. Though VLAD and IFK outperform BOVW, the main limitation is that they are of high dimension, resulting in high storage cost and low retrieval efficiency. The results of these mid-level features for each class are shown in Figure 6. Generally, VLAD is the best mid-level feature for most of the classes.

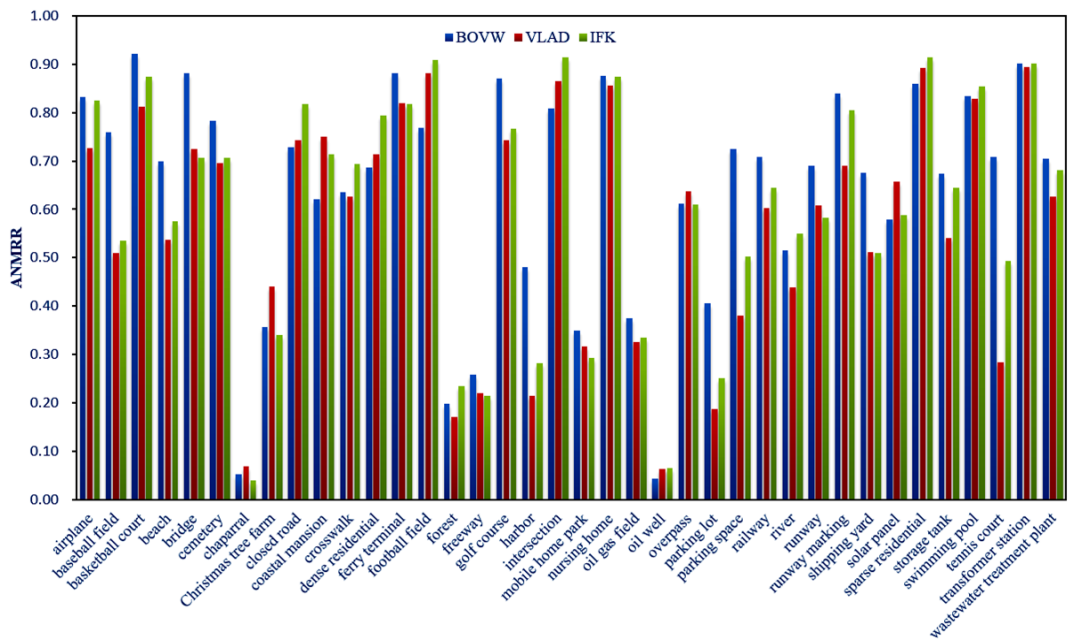

**Figure 6**. The results of mid-level features for each class in the PatternNet dataset. For BOVW representation, BOVW-128 is selected.

*5.2.3. Results of Deep Learning Features*

Table 4 shows the results of the deep learning feature based methods including the unsupervised feature learning method (UFL) and several pre-trained CNNs. For UFL features, we investigate the performance of UFL extracted with different numbers of neural units in the hidden layer. We can see UFL extracted with 400 hidden units performs better than the other UFL configurations. The pre-trained CNN features improve over the performance of UFL by more than 30% in terms of ANMRR values, indicating that supervised CNNs produce more discriminative features.

**Table 4.** The results of deep learning features on PatternNet. For ANMRR, lower values indicate better performance, while for mAP and *P@k,* larger are better. "UFL-K" means UFL with K neural units in the hidden layer.

| Features | ANMRR | mAP | P@5 | P@10 | P@50 | P@100 | P@1000 |
|---|---|---|---|---|---|---|---|
| UFL-400 | 0.6574 | 0.2525 | 0.5937 | 0.5646 | 0.4920 | 0.4516 | 0.2442 |
| UFL-600 | 0.6588 | 0.2508 | 0.5903 | 0.5629 | 0.4898 | 0.4497 | 0.2430 |
| UFL-800 | 0.6595 | 0.2501 | 0.5902 | 0.5619 | 0.4890 | 0.4489 | 0.2426 |
| AlexNet_Fc1 | 0.3328 | 0.6003 | 0.9545 | 0.9438 | 0.8986 | 0.8617 | 0.4934 |
| AlexNet_Fc2 | 0.3260 | 0.6042 | 0.9448 | 0.9331 | 0.8872 | 0.8529 | 0.4985 |
| CaffeRef_Fc1 | 0.3134 | 0.6221 | 0.9602 | 0.9511 | 0.9121 | 0.8787 | 0.5083 |
| CaffeRef_Fc2 | 0.3133 | 0.6171 | 0.9475 | 0.9370 | 0.8936 | 0.8604 | 0.5086 |
| VD16_Fc1 | 0.3302 | 0.6020 | 0.9388 | 0.9268 | 0.8806 | 0.8459 | 0.4959 |
| VD16_Fc2 | 0.3283 | 0.5986 | 0.9327 | 0.9204 | 0.8740 | 0.8404 | 0.4972 |
| VD19_Fc1 | 0.3423 | 0.5869 | 0.9352 | 0.9210 | 0.8694 | 0.8320 | 0.4865 |
| VD19_Fc2 | 0.3448 | 0.5789 | 0.9253 | 0.9113 | 0.8605 | 0.8247 | 0.4840 |
| VGGF_Fc1 | 0.3184 | 0.6170 | 0.9592 | 0.9493 | 0.9080 | 0.8738 | 0.5033 |
| VGGF_Fc2 | 0.3005 | 0.6309 | 0.9544 | 0.9442 | 0.9028 | 0.8714 | 0.5174 |
| VGGM_Fc1 | 0.3124 | 0.6231 | 0.9576 | 0.9472 | 0.9055 | 0.8717 | 0.5086 |
| VGGM_Fc2 | 0.3110 | 0.6188 | 0.9511 | 0.9405 | 0.8958 | 0.8627 | 0.5087 |
| VGGS_Fc1 | 0.3070 | 0.6290 | 0.9595 | 0.9508 | 0.9112 | 0.8784 | 0.5129 |
| VGGS_Fc2 | 0.2982 | 0.6333 | 0.9547 | 0.9449 | 0.9047 | 0.8734 | 0.5192 |
| GoogLeNet | 0.2983 | 0.6311 | 0.9445 | 0.9331 | 0.8918 | 0.8603 | 0.5202 |
| ResNet50 | **0.2606** | **0.6788** | **0.9665** | **0.9594** | **0.9274** | **0.9006** | **0.5533** |
| ResNet101 | 0.2624 | 0.6765 | 0.9638 | 0.9551 | 0.9208 | 0.8933 | 0.5525 |
| ResNet152 | 0.2632 | 0.6757 | 0.9635 | 0.9550 | 0.9208 | 0.8939 | 0.5511 |

The best performance of the various CNNs is achieved by ResNet50, showing that the deeper networks tend to achieve better performance than the shallower networks (i.e. AlexNet, CaffeRef, VGG, VGG-VD and GoogLeNet). However, the increased depth does reduce the performance when the network is too deep (see the performance of ResNet101 and ResNet152). It can also be observed that the features extracted from the second fully-connected layer (Fc2 feature) outperform the features extracted from the first fully-connected layer (Fc1 feature) except for the VD19 network. A possible explanation is that the second fully-connected layer is connected to the classifier layer and hence learns higher-level information. The results of these deep learning features for each class are shown in Figure 7.

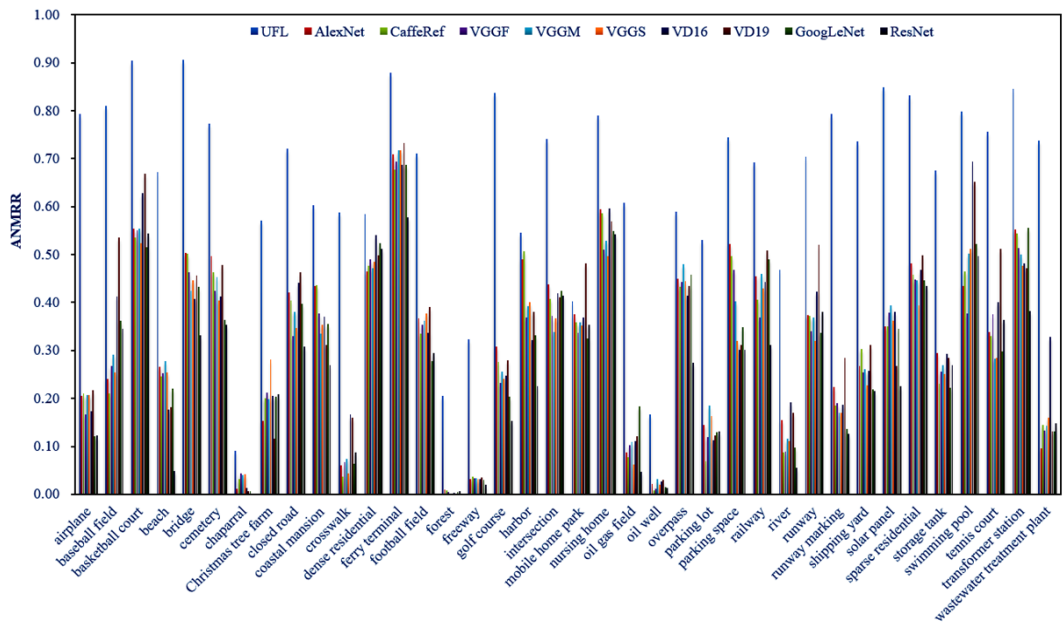

**Figure 7**. The results of deep learning features for each class in the PatternNet dataset.

Figure 8 shows the precision-recall curves for the handcrafted features and deep learning features. For families of features, the configuration that achieves the best performance is selected, namely BOVW-128, UFL-400, AlexNet_Fc2, CaffeRef_Fc2, VGGF_Fc2, VGGM_Fc2, VGGS_Fc2, VD16_Fc2, VD19_Fc1, and ResNet50.

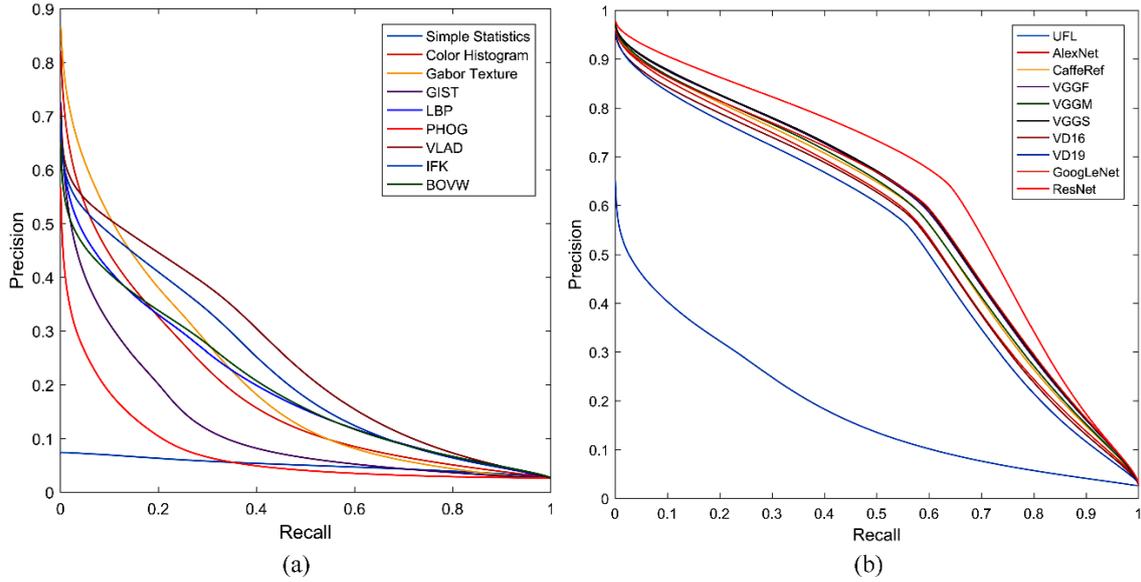

**Figure 8**. The precision-recall curves of handcrafted feature based methods and deep learning feature based methods: (a) precision-recall curves of handcrafted features, and (b) precision-recall curves of deep learning features.

Though the pre-trained CNNs achieve remarkable performance, their features are usually thousands of dimensions which are not compact enough for large-scale RSIR. In contrast, LDCNN is able to generate low-dimensional features. LDCNN is compared with handcrafted low-level and mid-level features, as well as deep learning features including UFL and some pre-trained CNNs on the 20% PatternNet images. As shown in Table 5, the results indicate that LDCNN outperform the pre-trained CNNs such as VGGF (the basic block of LDCNN), VGGS and even ResNet50 which achieves the best performance on PatternNet. The features extracted by LDCNN are 38-D which are pretty compact compared to the features extracted by the pre-trained CNNs.

**Table 5.** The comparisons of LDCNN and some other features. For ANMRR, lower values indicate better performance, while for mAP and *P@k,* larger is better. The handcrafted features and UFL are extracted under optimal configurations (i.e. the configurations that achieve the best performance on the entire PatternNet dataset).

| Features | ANMRR | mAP | P@5 | P@10 | P@50 | P@100 | P@1000 |
|---|---|---|---|---|---|---|---|
| Gabor Texture | 0.6439 | 0.2773 | 0.6855 | 0.6278 | 0.4461 | 0.3552 | 0.0899 |
| VLAD | 0.5677 | 0.3410 | 0.5825 | 0.5570 | 0.4757 | 0.4111 | 0.1104 |
| UFL | 0.6584 | 0.2535 | 0.5209 | 0.4882 | 0.3811 | 0.3192 | 0.0979 |
| VGGF_Fc1 | 0.3177 | 0.6195 | 0.9246 | 0.9037 | 0.7926 | 0.6905 | 0.1425 |
| VGGF_Fc2 | 0.2995 | 0.6337 | 0.9152 | 0.8964 | 0.7999 | 0.7047 | 0.1452 |
| VGGS_Fc1 | 0.3050 | 0.6328 | 0.9274 | 0.9070 | 0.8003 | 0.7013 | 0.1436 |
| VGGS_Fc2 | 0.2961 | 0.6374 | 0.9192 | 0.9009 | 0.8021 | 0.7073 | 0.1455 |
| ResNet50 | 0.2584 | 0.6823 | 0.9413 | 0.9241 | 0.8371 | 0.7493 | 0.1464 |
| LDCNN | 0.2416 | 0.6917 | 0.6681 | 0.6611 | 0.6747 | 0.6880 | 0.1408 |

## 6. Conclusions

We presented PatternNet, the largest publicly available remotely sensed evaluation dataset constructed for RSIR. We expect PatternNet help advance the state-of-the-art in RSIR, particularly deep learning based methods which require large amounts of labeled training data. We also surveyed a large number of RSIR approaches including traditional handcrafted features and recent deep learning features and evaluated them on PatternNet to establish baseline results to inform future research.